\documentclass[runningheads]{llncs}
\usepackage[]{graphicx}
\usepackage{url}
\usepackage[hidelinks]{hyperref}
\hypersetup{
    colorlinks = true,	
    citecolor = magenta, 
}

\usepackage{amsmath}
\usepackage{amssymb}
\usepackage{subcaption}
\usepackage{adjustbox}
\usepackage{algorithm}
\usepackage[]{algpseudocode}
\newcommand*{\Break}[1]{\textbf{break}{\ #1}}
\algblockdefx[Loop]{Loop}{EndLoop}[1][]{\textbf{loop} #1}{\textbf{end loop}}

\usepackage{xpatch}
\makeatletter
\xpatchcmd{\algorithmic}{\itemsep\z@}{\itemsep=1.0ex plus2pt}{}{}
\makeatother

\newcommand*{\pstc}{$PS^{2}C$}
\newcommand*{\paa}[1]{\bar{#1}}
\newcommand*{\sax}[1]{\hat{#1}}

\begin{document}
\title{Pattern Sampling for Shapelet-based \\ Time Series Classification
}
%
%


\author{%
Atif Raza\orcidID{0000-0003-1856-7286} \and
Stefan Kramer\orcidID{0000-0003-0136-2540}
}
\authorrunning{A. Raza and S. Kramer}
%
\institute{Johannes Gutenberg University Mainz, Staudingerweg 9, 55128 Mainz, Germany
\email{\{raza,\ kramer\}@informatik.uni-mainz.de}\\
\url{https://www.datamining.informatik.uni-mainz.de/}
}
\maketitle              
\begin{abstract}
Subsequence-based time series classification algorithms provide accurate and interpretable models, but training these models is extremely computation intensive.
The asymptotic time complexity of subsequence-based algorithms remains a higher-order polynomial, because these algorithms are based on  exhaustive search for highly discriminative subsequences.
Pattern sampling has been proposed as an effective alternative to mitigate the pattern explosion phenomenon.
Therefore, we employ pattern sampling to extract discriminative features from discretized time series data.
A weighted trie is created based on the discretized time series data to sample highly discriminative patterns.
These sampled patterns are used to identify the shapelets which are used to transform the time series classification problem into a feature-based classification problem.
Finally, a classification model can be trained using any off-the-shelf algorithm.
Creating a pattern sampler requires a small number of patterns to be evaluated compared to an exhaustive search as employed by previous approaches.
Compared to previously proposed algorithms, our approach requires considerably less computational and memory resources. Experiments demonstrate how the proposed approach fares in terms of classification accuracy and runtime performance.

\end{abstract}
\section{Introduction}
\label{intro}

The ubiquitousness of time series data implies that almost every human endeavor can benefit from time series data mining research, therefore, significant research efforts have been made in this regard over the past couple of decades.
Time series classification, specifically \textit{Shapelets} based time series classification, is a key research topic in the time series data mining domain \cite{Ye2011}. In contrast to deep neural network methods for time series, shapelet based methods can explicitly list relevant patterns and pattern occurrences used for classification, and thus belong to the category of methods that give explainable predictions in the first place. Formally, shapelets are subsequences that frequently occur in a specific class of time series instances while being absent or infrequent in the instances of the other classes.
Shapelet discovery is an exhaustive search process over all possible subsequences of a time series dataset, and the time required for shapelet discovery from a dataset with $N$ time series instances each of length $n$ is on the order of $O(N^2 n^4)$.

The Shapelet Transform (ST) algorithm extracts multiple shapelets in a single call to the shapelet discovery process and transforms the time series classification problem into a feature-based classification problem \cite{Hills2014:ShapeletTransform}.
Basically, an $N \times k$ dataset is created, where the rows and columns correspond to time series instances and shapelets, respectively, and each $(i,j)$ cell contains the minimum distance between the $i$th time series instance and the $j$th shapelet.
Subsequently, any off-the-shelf classification algorithm can be used for model induction using the feature set.
The evaluation step involves the calculation of $k$ distance values corresponding to the $k$ shapelets and using the induced model for classification.

Shapelet based time series classification for very large datasets requires a drastic reduction in the algorithmic complexity.
One way of addressing this issue is to transform the time series data into a symbolic representation.
The time series community has recognized and acknowledged the benefits of discretizing time series data \cite{Lin2012,Rakthanmanon2013,Senin2013,Schafer2015,Schafer2016,LeNguyen2019MrSEQL,Raza2020AcceleratingApproach}.
However, the approaches are still suffering from high computational complexity: the complexity of the Fast Shapelets (FS) \cite{Rakthanmanon2013} approach is $O(N n^2)$, the one of Bag of Patterns (BoP) \cite{Lin2012} is $O(N n^3)$, the one of Symbolic aggregate approximation - Vector Space Model (SAX-VSM) \cite{Senin2013} $O(N n^3)$, the one of Bag of SFA Symbols (BOSS) \cite{Schafer2015} $O(N^2 n^2)$, the one of Bag of SFA Symbols in Vector Space (BOSS VS) \cite{Schafer2016} is $O(N n^\frac{3}{2})$, and the complexity of Mr-SEQL \cite{LeNguyen2019MrSEQL} is $O(N n^{\frac{3}{2}} \mathrm{log}\ n)$.\footnote{Notice that another well-known algorithm, the Matrix Profile \cite{Yeh2016}, is not applicable in this setting, because it takes a single, long time series as an input.}
A recent approach called MiSTiCl directly employs string mining for frequent pattern extraction from discretized time series datasets \cite{Raza2020AcceleratingApproach} and has a complexity on the order of only $O(Nnl)$.
The authors of MiSTiCl noted that the pattern extraction phase consumes approximately 80\% of the total time although the string mining algorithm used as the pattern extractor has a linear time complexity in the length of all discretized time series instances concatenated \cite{Raza2020AcceleratingApproach}.
This can be attributed to the pattern explosion problem when searching for frequent patterns, since the number of possible subsequences in an $m$ character long string based on an alphabet size $\alpha$ is $\frac{1-\alpha^{m+1}}{1-\alpha}$.

Basic pattern mining involves enumerating all possible pattern combinations to find interesting patterns, but this results in the infamous pattern explosion problem.
Different approaches have been proposed to address this phenomenon, however, these approaches have their own associated drawbacks, e.g., related and redundant patterns lacking diversity, too few or too many patterns, high computational cost, etc.
Pattern sampling is an alternative to the exhaustive pattern enumeration approach, and a number of variations of pattern sampling have been proposed.
Pattern sampling proposes to sample one pattern at a time proportional to a quality measure \cite{Dzyuba2017FlexibleMining}.
The aim is to limit the number of patterns that are evaluated upfront, but still being able to evaluate additional patterns when the need arises or the consumer process intends to evaluate more patterns with an aim to improve the overall accuracy.

In this paper, we propose the first pattern sampling approach for shapelet based time series classification.
We use a discretized representation of the time series data for shapelet discovery and replace the exhaustive frequent pattern extraction step with a pattern sampler.\footnote{We refer to real-valued time series segments as ``subsequences'' and the discretized/symbolic segments as ``patterns''.}
Our approach provides competitive accuracy compared to state-of-the-art pattern based time series classification approaches and has an on par computational complexity as the most efficient pattern based time series classification approaches known today.
In the following, our approach will be referred to as \underline{P}attern \underline{S}ampling for \underline{S}eries \underline{C}lassification (\textit{\pstc{}}).

\section{Pattern Sampling for Time Series Classification}
\label{sec:algo}

Our proposed algorithm stands out from other pattern based time series classification algorithms, because it employs a pattern sampler instead of evaluating all the candidate patterns for finding the most discriminative shapelets.
The basic structure of our algorithm is similar to other feature/pattern based time series classification algorithms, e.g., ST \cite{Hills2014:ShapeletTransform}, MiSTiCl \cite{Raza2020AcceleratingApproach}, etc.
The main steps of the algorithm are: (i) discretizing the time series data, (ii) creating a pattern sampler, (iii) creating a feature set via sampling a fixed number of patterns or until a quality threshold is met, (iv) creating a transformed dataset using the sampled patterns, and finally (v) model induction.

\begin{figure}[t]
    \centering
    \includegraphics[width=0.9\textwidth]{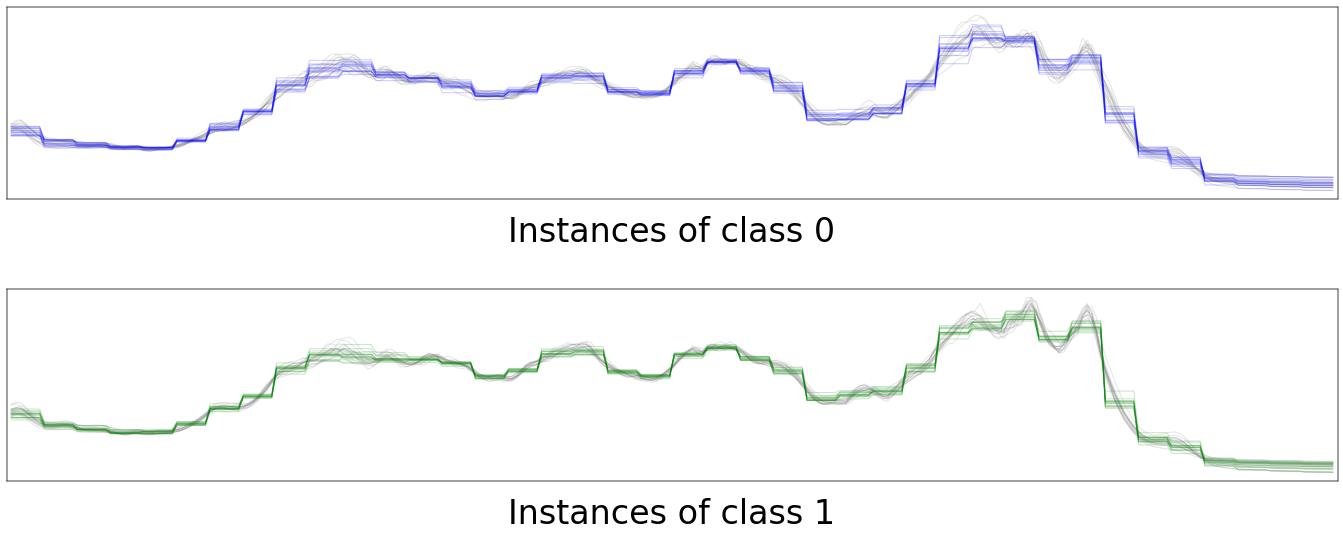}    
    \caption[PAA version of time series instances superimposed on their real-valued counterparts]
            {PAA version of time series instances superimposed on their real-valued counterparts. The original time series instances (shown in light gray) have 286 data points, and the PAA versions have 40 data points based on a dimensionality reduction factor $\omega=7$. The PAA versions have been stretched (along the x-axis) to emphasize the retention of the overall shape of time series instances.}
    \label{fig:real-and-saxed}
\end{figure}

A \textit{time series} is an ordered, real-valued sequence of $n$ observations denoted as $T=(t_1,t_2,\ldots,t_n)$.
A label $y \in C$ can be assigned to a time series instance, where $C$ is the set of all class labels.
A time series dataset $D$ consists of $N$ labeled time series instances $\{(T_1,y_1),(T_2,y_2),\dots,(T_N,y_N)\}$.
%
Symbolic aggregate approximation (SAX) is a widely used time series discretization algorithm \cite{Lin2007a}.
It transforms a time series $T$ of length $n$ into a string $\sax{T}=(\sax{t}_1, \sax{t}_2,\ldots,\sax{t}_p)$ of length $p=\lfloor\frac{n}{\omega}\rceil$, where $p~\ll~n$ and $\omega$ represents the averaging window size.
Each non-overlapping sequence of $\omega$ observations of $T$ is averaged to provide one observation, i.e., $\paa{T} = (\paa{t}_1,\paa{t}_2,\ldots,\paa{t}_p)$.
Figure \ref{fig:real-and-saxed} illustrates real-valued time series instances and their PAA versions.
Next, each observation $\paa{t}_i \in \paa{T}$ is mapped to a character from an alphabet of size $\alpha \in \mathbb{Z}_{\geq 2}$ such that $\sax{t}_i = \mathnormal{alpha}_j,\ \mathrm{iff}\ \beta_{j-1} \leq \paa{t}_i < \beta_j$.
The quantization blocks for the alphabet are chosen based on breakpoints $(\beta_0,\beta_1,\ldots,\beta_{\alpha})$, where $\beta_0$ and $\beta_{\alpha}$ are defined as $-\infty$ and $\infty$, respectively, and the remaining breakpoints are chosen such that area under the $\mathcal{N}(0,1)$ Gaussian curve from $\beta_i$ to $\beta_{i+1}$ equals $\frac{1}{\alpha}$.

Discretizing time series data using SAX preserves the overall shape, however, it can also lead to a loss of temporal features.
This is an artifact of inadvertent feature splitting due to the use of an arbitrary window size $\omega$.
One effective way of dealing with this problem is to initiate multiple independent feature extraction pipelines, each based on a different combination of $\alpha$ and $\omega$, and finally merging the results of each individual feature extraction problem into one aggregate transformed dataset.
This results in a diverse feature set that leads to better overall accuracy when using an ensemble classifier for model induction, because ensemble methods inherently tend to reduce variance and sometimes also bias.
This multi-resolution feature extraction technique has been effectively used in a number of pattern-based time series classification approaches \cite{Schafer2015,Raza2020AcceleratingApproach,LeNguyen2019MrSEQL}.

Algorithm \ref{alg:main-algo} lists the main steps of the \pstc{} algorithm. 
The first step is the discretization of the train and test splits corresponding to the current $(\alpha, \omega)$ parameter combination (Line~\ref{alg:main-algo:discretize-splits}).
Next, the discretized training set instances are used to create suffix tree representations for fast substring searching (Line~\ref{alg:main-algo:preprocess-trainset}).
The choice of suffix trees is rather superficial, since there are a number of other data structures that can be utilized to efficiently search a given string for the presence of query substrings.
Next, a probabilistic pattern sampler is induced based on the discretized training data (Line~\ref{alg:main-algo:extract-freq-patts}).
Next, transformed training and testing feature sets are created based on sampled patterns (Line~\ref{alg:main-algo:populate-splits}).
Once all individual feature sets have been created, they are concatenated to form a single feature set, which can be used to induce a classification model (Line~\ref{alg:main-algo:optimization-step}).
\begin{algorithm}[t]
    \caption{\pstc{}($D_{train}$, $D_{test}$, $A$, $\Omega$, $l_{max}$, $s_{min}$, $\tau$, $K$)}
    \label{alg:main-algo}
    \begin{algorithmic}[1]
        
        
        \State $SV_{train} \gets \{\}$, $SV_{test} \gets \{\}$
        \Comment{Initialize associative arrays for feature sets}
        \label{alg:main-algo:initialize-feature-set-maps}
        
        \ForAll{$(\alpha,\omega) \in A \times \Omega$}
        \label{alg:main-algo:alpha-w-loop-start}
        
            \State $\widehat{D}_{train}, \widehat{D}_{test} \gets $\Call{Discretize}{$D_{train}$, $D_{test}$, $\alpha$, $\omega$}
            \label{alg:main-algo:discretize-splits}
            
            \State $\widehat{D'}_{train} \gets $\Call{CreateSuffixTrees}{$\widehat{D}_{train}$}
            \label{alg:main-algo:preprocess-trainset}
            
            \State $Sampler \gets $\Call{FitPatternSampler}{$\widehat{D}_{train}$, $\widehat{D'}_{train}$, $l_{max}$, $s_{min}$, $\tau$}
            \label{alg:main-algo:extract-freq-patts}
            
            \State $SV_{train}^{\alpha,\omega}, SV_{test}^{\alpha,\omega} \gets $\Call{CreateFeatureSet}{$D_{train}$, $D_{test}$, $\widehat{D}_{train}$, $Sampler$, $K$}
            \label{alg:main-algo:populate-splits}
        
        \EndFor
        \label{alg:main-algo:alpha-w-loop-end}
        
        \State $FS_{train}, FS_{test} \gets$ \Call{MergeFeatureSets}{$SV_{train}$, $SV_{test}$, $A$, $\Omega$, $|D_{train}|$}
        \label{alg:main-algo:optimization-step}
        
        \State \Return $FS_{train}, FS_{test}$
    \end{algorithmic}
\end{algorithm}

\subsection{Creating the Pattern Sampler}
\label{subsec:patt-samp}

A pattern sampler can be modeled in several ways, e.g., graphs (MCMC), trees, etc. \cite{Dzyuba2017FlexibleMining}.
We have envisaged the pattern sampler as a trie with weighted edges,  since this allows to incorporate constraints and fast, iterative updates to the sampler.
A trie is a data structure used to store strings in order to support fast pattern matching.
Formally, if $S$ is a set of $s$ strings from an alphabet $\Sigma$, then a standard trie for $S$ is an ordered tree with the following properties:
\begin{itemize}
    \item Each node of a trie, except the root, is labeled with a character of $\Sigma$.
    \item The children of an internal node of the trie have distinct labels.
    \item The trie has $s$ leaves, each associated with a string of $S$, such that the concatenation of the labels of the nodes on the path from the root to a leaf $v$ of the trie yields the string of $S$ associated with $v$.
\end{itemize}
Thus, a trie represents the strings of $S$ with paths from the root to the leaves.
For strings sharing a common prefix, the edges are shared for the common prefixes and a split is created when the characters in the strings differ.

In addition, we augment the trie with weighted edges such that inserting a string in the trie also associates a corresponding weight to all the inserted edges.
An edge shared between multiple strings has a weight equal to the aggregate of the weights associated with all the strings that share the particular edge.
%
Edge weights are based on the discriminative capability of inserted patterns.
The $\chi^2$ statistic can be used to determine whether there is a statistically significant difference between the expected and observed counts for a given contingency table consisting of two or more categories.
In case of a symbolic time series dataset, the categories are the different classes, while the counts are the number of instances belonging to each class in which the given pattern is present or absent.
The range of values for the $\chi^2$ statistic is $(0, |D_{train}|]$.
For a binary class problem, if a pattern occurs in all the instances of one class, whereas it is absent in all instances of the other class, then the $\chi^2$ statistic will be maximized, whereas if the pattern is present/absent in most of the instances, then the $\chi^2$ statistic will be close to 0.
In order to simplify subsequent steps, the $\chi^2$ statistic is normalized with $|D_{train}|$ so that the effective range becomes $(0, 1]$, where the value of 1 indicates that the given pattern is a perfect discriminator, while a value close to 0 indicates otherwise.

\begin{figure}[t]
    \centering
    \includegraphics[width=0.85\textwidth]{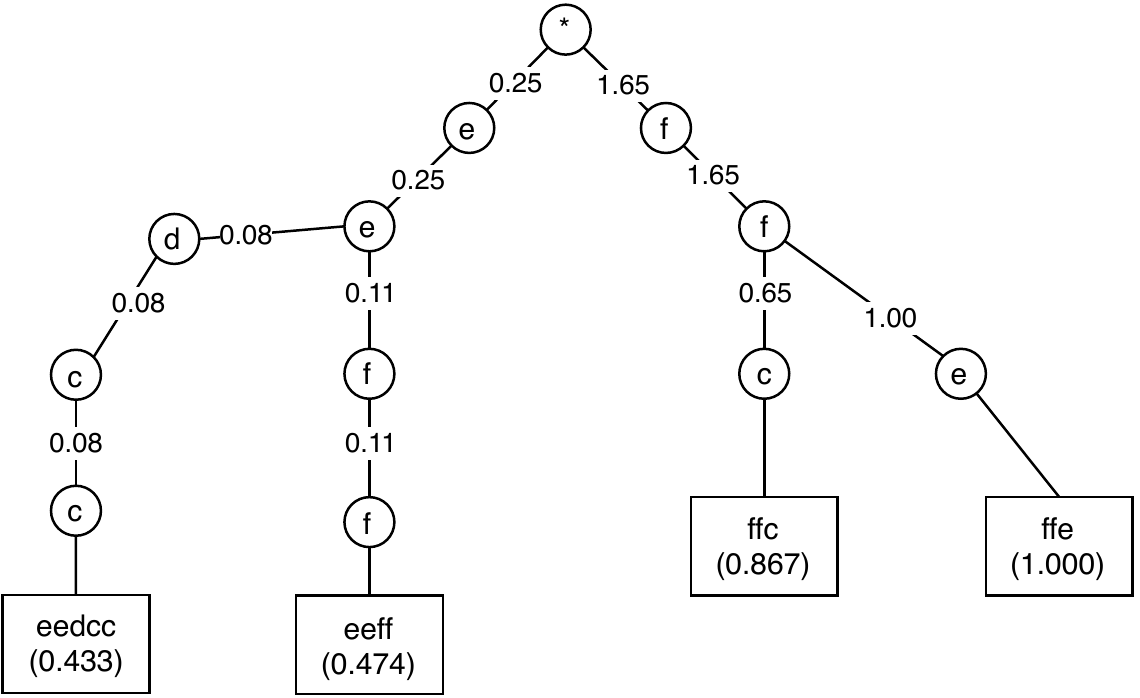}
    \caption{An illustration of a weighted trie. The edges are weighted using the scaled quality measures of the strings. The weight of a shared edge is equal to the aggregated weights contributed by all strings sharing that edge. The leaf nodes show the patterns added to the trie along with their quality measures.}
    \label{fig:example-trie}
\end{figure}

The normalized $\chi^2$ statistics can be directly used as weights for the edges, however, we can introduce a bias towards highly discriminative patterns using temperature scaling.
The scaled edge weights are calculated as $q^{(1/\tau)}$, where $q$ is the normalized $\chi^2$ statistic and $\tau$ is the temperature scaling factor.
During pattern sampling, the probability of selecting an edge is given as $f_{\tau}(q)_{i} = \frac{q_{i}^{(1/\tau)}}{\sum_{j} q_{j}^{(1/\tau)}}$, where $\sum_{j} q_{j}^{(1/\tau)}$ is the sum of all edge weights originating from the node.
When $\tau~=~1$, the edge weights are linearly proportional to the normalized $\chi^2$ statistics.
As $\tau$ decreases, the bias towards patterns with higher normalized $\chi^2$ statistics increases, e.g., a quadratic scaling is applied to the values for $\tau = 0.5$.
As $\tau \to 0$, the function turns into an argmax function.
Figure \ref{fig:example-trie} shows an example trie created from a set of words extracted from a discretized dataset.
The figure is based on the popular \textit{Coffee} dataset that is a binary class dataset with 14 instances in each class.
The scaling factor $\tau$ is set to 0.33, alphabet size $\alpha$ is set to 6, and dimensionality reduction factor $\omega$ is set to 4.
The pattern \texttt{ffe} occurs in all instances of one class and has a normalized $\chi^2$ statistic of 1.0 that translates into a scaled edge weight of 1.0.
Another pattern \texttt{ffc} occurs in 13 of the 14 instances of the other class, therefore, its normalized $\chi^2$ statistic is equal to 0.867 and the scaled weight is equal to 0.65.
Inserting the first pattern adds the required edges with each associated edge weights.
When the second pattern is inserted, the edges corresponding to substring \texttt{ff} have their weight updated to be the sum of the previous weight and the weight associated with the current pattern, while a new edge is inserted for the suffix \texttt{c} with the respective weight for the pattern.
The other patterns are also inserted similarly.

Algorithm \ref{alg:trie-algo} lists the steps involved in the creation of a weighted trie based on patterns up to a user specified length.
The procedure extracts all patterns of a given length using the suffix trees.
Next, each candidate pattern is evaluated to determine its discriminative capability using the $\chi^2$ statistic.
If the normalized $\chi^2$ statistic is greater than or equal to $s_{min}$, the pattern is accepted to be inserted in the trie, otherwise it is discarded.
Starting from the root node, the insertion procedure checks if an edge corresponding to the first character in the candidate pattern is present or not.
If the edge is absent, the procedure adds the edge and sets the edge weight equal to the scaled quality measure for the pattern.
If an edge corresponding to the character is already present, then the edge weight is updated by adding the scaled quality measure of the pattern.
Similarly, the node weight is updated according to the new aggregate of edge weights.
The procedure then traverses down the edge added/updated and checks for the second character in the pattern and so on, until all the characters have been inserted.
\begin{algorithm}[t]
    \caption{FitPatternSampler ($\widehat{D}_{train}$, $\widehat{D'}_{train}$, $l_{max}$, $s_{min}$, $\tau$)}
    \label{alg:trie-algo}
    \begin{algorithmic}[1]

        
        \State Initialize an empty trie $Sampler$
        
        \For{$l \gets 2$ to $l_{max}$}
        \State $S \gets $\Call{FindAllPatternsWithLength}{$l, \widehat{D'}_{train}$}
            \For{$s \in S$}
            
                \State $q \gets $\Call{CalculateChiSqStatistic}{$s,\widehat{D}_{train}, \widehat{D'}_{train}$}
                \If{$q \geq s_{min}$}
                    \State \Call{Insert}{$s,q,\tau$}
                \EndIf
                
            \EndFor

        \EndFor

        \State \Return $Sampler$
    \end{algorithmic}
\end{algorithm}

\subsection{Creating Feature Sets}
\label{subsec:algo-creating-feature-sets}

The next step is the creation of a feature based dataset with $K$ sampled patterns.
Algorithm \ref{alg:transform} lists the pseudo-code for creating the real-valued feature datasets.
After initialization of the necessary data structures, $K$ patterns are sampled from the trie.
Sampling a pattern involves traversing the weighted trie from the root node to a leaf node using the fitness proportionate (roulette wheel) selection method.
At any node, the probability of selecting the $i$th edge is calculated by dividing the edge weight $q_i^{1/\tau}$ by the sum of all edge weights for the current node $\sum_j q_j^{1/\tau}$.
A uniformly distributed random number $r$ is drawn in the range $\left[0,\sum_j q_j^{1/\tau}\right)$.
Now, the edge weights of all edges are compared with the random number $r$ in their lexical order.
For each edge $i$, if $r$ is less than the edge weight $q_i^{1/\tau}$ then the $i$th edge is selected as the next edge, otherwise $q_i^{1/\tau}$ is subtracted from $r$ and the next edge weight is compared.
The process continues until the last edge originating from the node.
If a node has child nodes and is also a leaf node, the decision to return the string terminating at the current node or to traverse the trie further is also based on a random number.

For each sampled pattern, a reverse lookup is performed to get the real-valued subsequences from the symbolic patterns.
Next, $K$-column feature sets are created, for the training and test set, respectively.
The rows of the feature sets correspond to the time series instances, and columns represent the $K$ shapelets discovered.
The cells are populated with the distance values between the time series instances and the discovered shapelets.
\begin{algorithm}[t]
    \caption{CreateFeatureSets ($D_{train}$, $D_{test}$, $\widehat{D}_{train}$, $Sampler$, $K$)}
    \label{alg:transform}
    \begin{algorithmic}[1]
        
        
        \State $SV_{train} \gets MATRIX(|D_{train}|, K)$, $SV_{test} \gets MATRIX(|D_{test}|, K)$\label{alg:transform:init-train-test-sv}
        
    	\For{$k \gets 1$ to $K$}
        \label{alg:transform:f-loop-start}
            
            \State $f \gets $\Call{SamplePattern}{$Sampler$}
            
        	\State $s \gets $\Call{PerformReverseLookup}{$D_{train}$, $\widehat{D}_{train}$, $f$}
            \label{alg:transform:perform-rev-lookup}
            
            \State \begin{varwidth}[t]{\linewidth}
            			For each $T \in D_{train}$\par
                        \hskip\algorithmicindent populate the respective row and column of $SV_{train}$ \par
                        \hskip\algorithmicindent with the distance value between $T$ and $s$
                  \end{varwidth}
            \label{alg:transform:train-set-inst}
            
            \State \begin{varwidth}[t]{\linewidth}
            			For each $T \in D_{test}$\par
                        \hskip\algorithmicindent populate the respective row and column of $SV_{test}$ \par
                        \hskip\algorithmicindent with the distance value between $T$ and $s$
                  \end{varwidth}
			\label{alg:transform:test-set-inst}
            
            \State Increment $k$ and if $k$ equals $K$ \Break{Outer loop}
		
        \EndFor
        \label{alg:transform:f-loop-end}

        \State \Return $SV_{train}, SV_{test}$
    \end{algorithmic}
\end{algorithm}

\subsection{Merging Individual Feature Set}

Combining the feature sets created for each $(\alpha, \omega) \in A \times \Omega$ parameter combination yields a feature set which can mitigate the problem of feature loss due to discretization while providing increased accuracy with the inclusion of features obtained for different levels or resolutions of discretization and quantization.
Since use of multi-resolution feature sets has been inspected previously, there have also been attempts at optimizing the overall results by merging only the feature sets which can contribute the most towards improving the classification accuracy \cite{Raza2020AcceleratingApproach}.
In our experiments, it was observed that creating an optimized version of the merged feature set only provides a minor improvement in accuracy, if any.
In most cases, classification models created using a merged feature set without any optimization towards finding a perfect combination of individual feature sets are as accurate as the optimized feature set based classification models.

\subsection{Complexity Analysis}
\label{subsec:complexity}

The computational complexity can be determined by investigating a single feature set creation iteration based on an arbitrary combination of $\alpha$ and $\omega$.
SAX requires $O(Nn)$ operations to discretize a dataset.
The time taken for creating a pattern sampler depends on: (i) the time taken to extract candidate patterns, and (ii) the time taken in finding the candidate pattern in each discretized instance of the training set.
Therefore, the time required for both these steps is $O(Nm)$.
Sampling $K$ patterns is proportional to the maximum pattern length in the trie $O(l_{max})$.
For each feature, $N$ feature values have to be calculated, where each feature value calculation takes $O(ns)$ time, where $s$ is the length of a subsequence and $s \ll n$.
Since $K$ is a constant and much smaller than $N$ and $n$, the time required for creating a feature set is on the order of $O(Nns)$.
The overall time complexity of creating a feature set for a given $\alpha$ and $\omega$ parameter combination is on the order of $O(Nn)+O(N\frac{n}{w})+O(1)+O(Nns) \approx O(Nns)$.
Since the quantity $|A| \times |\Omega|$ is also constant, the asymptotic time complexity of the algorithm is on the order of $O(Nns)$.

\section{Empirical Evaluation}
\label{sec:experiments}

The UCR/UEA Time Series classifiction Repository\footnote{UCR/UEA Time Series Repository \url{https://www.timeseriesclassification.com}} has evaluated many time series classification algorithms using an extensive set of datasets and provides the classification accuracy results for comparison.
These results are based on 100 evaluations of each dataset using shuffled training and testing set splits.
This evaluation strategy has become a \textit{de facto} convention for reporting time series classification results.
We have also evaluated \pstc{} using the same evaluation strategy.
In order to compare \pstc{} against other well-known algorithms regarding classification accuracy, we have used the results provided by the UCR/UEA Repository and the repositories for the MiSTiCl and Mr-SEQL algorithms.
The runtime requirements for MiSTiCl, BOSS, BoP, and SAX-VSM were taken from the MiSTiCl repository.

All experiments were performed with a fixed set of parameters for all datasets.
The $A$ and $\Omega$ parameters were set to $\{2, 3, \ldots, 8\}$ and $\{2, 3, \ldots, 6\}$, respectively.
The maximum allowed pattern length $l_{max}$ was set to 20, the minimum acceptable discriminative power $s_{min}$ (normalized $\chi^2$ statistic) was set to be 0.05, the scaling factor $\tau$ was set to 0.5, and $K$ was set to 4.
For statistical comparison of different algorithms, we employ the Friedman test followed by Nemenyi post-hoc test based on average ranks attained by the different algorithms and show the comparisons as critical difference (CD) diagrams \cite{Demsar2006}.
The executable code and required scripts are available online.\footnote{Executable code and scripts available at: \url{https://drive.google.com/drive/folders/16oUBQ8ycGOwXnSgIsBRuTnl5MFuf2g4H?usp=sharing}.}


\subsection{Results}
\label{sec:results}

In terms of classification accuracy, our algorithm performs on par with other algorithms for datasets with two to six classes, however, classification accuracy deteriorates as the number of classes in a dataset goes beyond eight.
This behavior is due to the fact that we create a single pattern sampler and there is no provision for sampling class-correlated patterns.
An obvious alternative is to create samplers for each class individually in an one-vs-all fashion, however, another alternative is to incorporate additional information with the patterns, which could enable class-correlated pattern sampling.
The minimum acceptable discriminative power $s_{min}$ for a candidate pattern allows to adjust the acceptance threshold for candidate patterns.
A high value allows to accept only the very best patterns,  while a value close to zero allows to accept almost all patterns.
Accepting a large number of patterns can lead to a densely populated trie, but a stringent scaling factor $\tau$ can help deal in this case by heavily weighting the useful patterns and diminishing the chances of sampling less useful patterns.
Therefore, $s_{min}$ and $\tau$ are complementary parameters.
The maximum allowed pattern length $l_{max}$ is basically used to limit the number of patterns inserted into the trie.
In most cases, the discriminative patterns are much shorter than the length of discretized time series instances, however, many discriminative patterns can have a huge number of variants with either a prefix or a suffix.
The $l_{max}$ parameter allows to restrict the inclusion of too many variant patterns in the trie, and in doing so, helps to keep the trie balanced since the inclusion of too many variants with the same discriminative power would cause the sampling procedure to return related and/or redundant patterns.

\begin{figure}[t]
    \centering
    \includegraphics[width=0.75\textwidth]{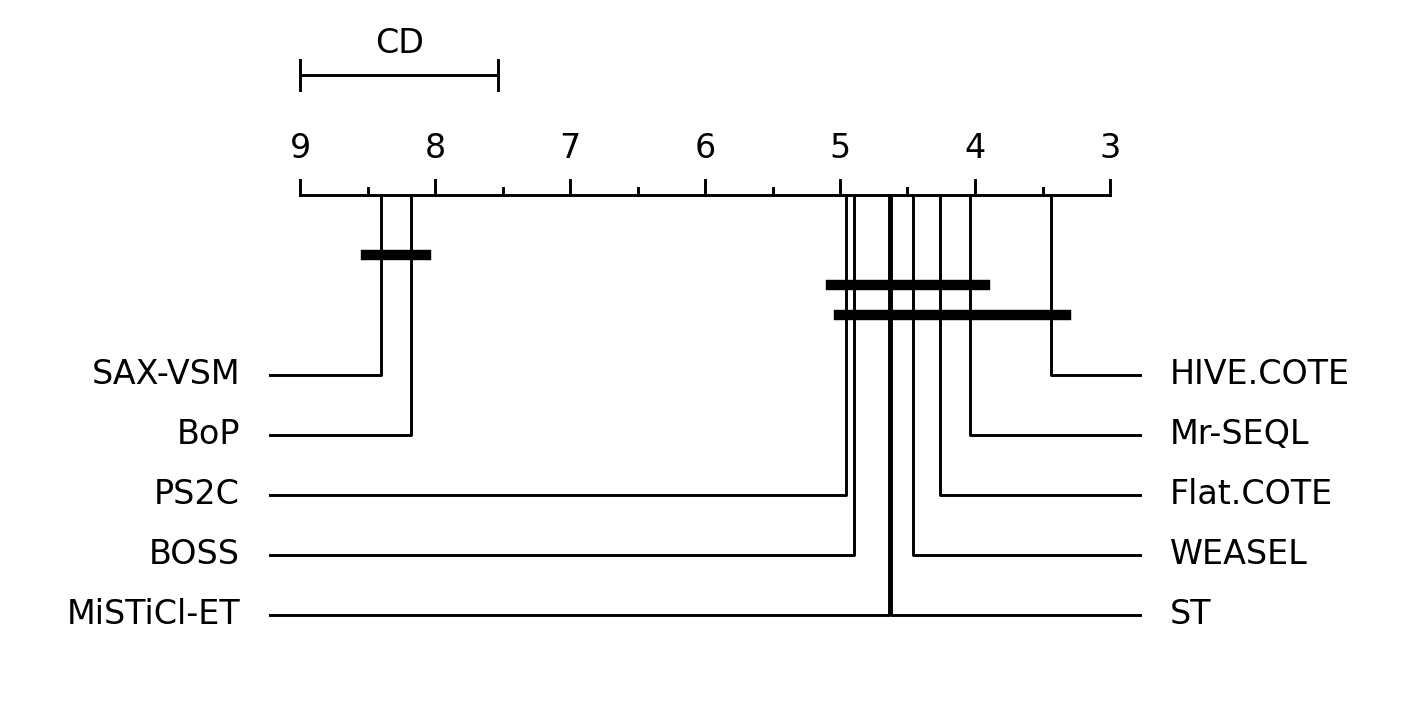}
    \caption{Average ranks based on classification accuracy for different time series classification algorithms. The critical difference (CD) for significantly different algorithms is 1.46.}
    \label{fig1}
\end{figure}

Figure \ref{fig1} shows a critical differences diagram for different time series classification algorithms regarding classification accuracy.
Overall, \pstc{} performs impressively and is on par with algorithms like ST and Flat.COTE.
HIVE.COTE and Flat.COTE are two ensemble classifiers which base their classification on the basis of various types of classifiers, including ST, BOSS, etc.
Both these algorithms have an extremely high computational cost due to their dependence on training several different types of classification algorithms.
Among the pattern based time series classification algorithms, SAX-VSM and BoP perform worse, while MiSTiCl, Mr-SEQL, BOSS and \pstc{} perform similarly and are not significantly different from each other.
\pstc{} is not significantly different from ST or Flat.COTE, however, it narrowly misses the group that forms the cohort of best performing time series algorithms in this comparison.

\begin{figure}[t]
    \centering
    \includegraphics[width=0.75\textwidth]{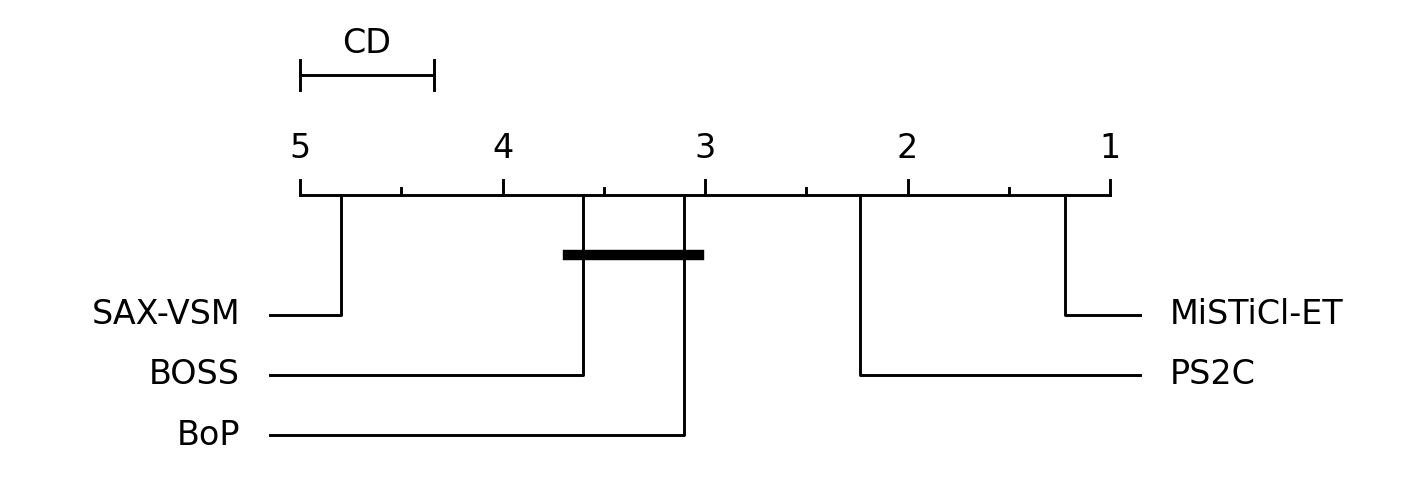}
    \caption{Average ranks based on runtime performance for \pstc{}, MiSTiCl, BOSS, BoP, and SAX-VSM. The critical difference (CD) for significantly different algorithms is 0.6x.}
    \label{fig2}
\end{figure}

Figure \ref{fig2} shows the critical differences diagram for pattern-based time series classification algorithms regarding running times.
MiSTiCl was a clear winner and \pstc{} was the second fastest, while BoP/BOSS and SAX-VSM were significantly slower than either of the two algorithms.
A direct comparison with other algorithms was not possible due to the lack of availability of runtime performance data from the UEA and Mr-SEQL repositories.
Overall, \pstc{} was 1.1 to 1.3 times slower than MiSTiCl on average, however, since MiSTiCl was shown to be significantly faster than the other algorithms, we can confidently assume that \pstc{} is also substantially faster than the remaining algorithms. This is backed up by complexity considerations (see Section 1 and Section 2.4).

\section{Conclusion}\label{sec:conclusions}

The paper introduced the first pattern sampling algorithm for time series data. The pattern sampler is used in a shapelet based classification algorithm. It was demonstrated that pattern sampling can be an effective alternative to the exhaustive shapelet/pattern discovery processes, since it enables to extract frequent patterns based on a quality measure to counteract the pattern explosion phenomenon.
We used a multi-resolution feature set creation approach in our experiments, since it is proven to be highly effective.
Our pattern sampling based algorithm was mostly on par with other similarly structured algorithms regarding classification accuracy.
In terms of computational costs, our approach is slightly slower than MiSTiCl, however, the complexity analysis indicates the asymptotic complexity for our approach is similar to that of MiSTiCl, implying that the proposed method is faster than the other algorithms.

Shapelet based time series classification gives rise to explainable classifications by construction. Therefore, the proposed pattern sampler is another option for constructing interpretable feature sets for time series. Interesting combinations with deep neural networks, especially for smaller sized datasets, remain a topic for future research \cite{Kramer2020}.

There are a few optimizations that have been identified as further future avenues to be explored.
We need to explore class-correlated pattern sampling in order to improve the accuracy in cases where the pattern sampler keeps providing patterns for one or a few classes rather than for the majority of classes.
We can also experiment with fuzzy pattern sampling to diversify the identified feature set per pattern sampler.

%
%
%
\bibliographystyle{splncs04}
\bibliography{references}
\end{document}